\documentclass[11pt]{article}

\usepackage{emnlp2023}

\usepackage{times}
\usepackage{latexsym}
\usepackage[T1,T5]{fontenc}

\usepackage[utf8]{inputenc}
\usepackage{microtype}
\usepackage{inconsolata}

\usepackage{graphicx}

\title{Recipe for Zero-shot POS Tagging: Is It Useful in Realistic Scenarios?}

\author{
 \textbf{Zeno Vandenbulcke},
 \textbf{Lukas Vermeire},
 \textbf{Miryam de Lhoneux}
\\
Department of Computer Science,
\\
 KU Leuven
\\
 \texttt{
    \href{mailto:zeno.vandenbulcke@student.kuleuven.be,lukas.vermeire@student.kuleuven.be}
    {\{zeno.vandenbulcke, lukas.vermeire\}@student.kuleuven.be},
    \href{mailto:miryam.delhoneux@kuleuven.be}
   {miryam.delhoneux@kuleuven.be}
}
}

\begin{document}
\maketitle
\begin{abstract}
POS tagging plays a fundamental role in numerous applications. While POS taggers are highly accurate in well-resourced settings, they lag behind in cases of limited or missing training data. This paper focuses on POS tagging for languages with limited data. We seek to identify the characteristics of datasets that make them favourable for training POS tagging models without using any labelled training data from the target language. This is a zero-shot approach. We compare the accuracies of a multilingual large language model (mBERT) fine-tuned on one or more languages related to the target language. Additionally, we compare these results with models trained directly on the target language itself. We do this for three target low-resource languages.
Our research highlights the importance of accurate dataset selection for effective zero-shot POS tagging. Particularly, a strong linguistic relationship and high-quality datasets ensure optimal results. For extremely low-resource languages, zero-shot models prove to be a viable option.
\end{abstract}

\section{Introduction}

In recent years, a lot of progress has been made in Natural Language Processing (NLP). However, certain fundamental technologies such as \emph{Part-of-Speech} (POS) \emph{tagging} or dependency parsing are still only available for a small part of the world's languages. This is mostly for languages with significant amounts of available data. For languages with limited or no available data (\textit{low-resource} languages), these technologies are highly inaccurate or sometimes even nonexistent \citep{joshi-etal-2020-state}. Advancements in multilingual language models have shown impressive cross-lingual transfer abilities \citep{wu-dredze-2019-beto}. In this paper, we build on these advancements to explore zero-shot POS tagging for low-resource languages.

\newpage

We investigate two questions: 
\begin{itemize}
    \item[\textbf{RQ1}] What are the essential characteristics of datasets for effectively fine-tuning \textit{zero-shot POS tagging models} for low-resource languages?
    \item[\textbf{RQ2}] Are zero-shot models useful in realistic low-resource settings when compared to models fine-tuned with target language data?
\end{itemize}

We explore these questions by fine-tuning a multilingual pretrained language model for zero-shot POS tagging, using related languages (which we call \emph{support languages}) to the target language.
We start by fine-tuning POS tagging models for Afrikaans, using Dutch, German, and English as support languages. We test the models on Afrikaans and compare the results in an attempt to identify the characteristics of the datasets that affect the performance of the models. We then experiment with two additional target languages: Faroese (supported by Icelandic, Danish, Norwegian and Swedish) and Upper Sorbian (supported by Czech, Polish and Slovak). We aim to determine whether our findings for Afrikaans also apply to these languages.

In relation to \textbf{RQ1}, we find that when multiple supporting languages are available, high-quality datasets \citep{kulmizev-nivre-2023-investigating} that are closely related to the target language result in better performance. Using the most closely related language leads to consistently better accuracy, especially with a limited number of training sentences. For an optimal training dataset size, using between 100 and 5000 sentences helps to avoid \textit{under-} or \textit{overfitting}.

Regarding \textbf{RQ2}, we find that zero-shot POS tagging models can certainly be a viable option for low-resource languages. Nevertheless, models trained on annotated data from the low-resource target language itself remain superior, similarly to previous findings in the literature \citep{meechan-maddon-nivre-2019-parse}. 

As we will discuss in the \hyperref[sec:conclusion]{conclusion}, these findings can be translated into concrete guidelines for different scenarios.

This paper starts with background information on low-resource zero-shot POS tagging. We then discuss the technical components and methodology used. Finally, we present the results and attempt to answer the previously mentioned research questions.

\section{Background}
\label{sec:background}
\subsection*{Part-of-Speech tagging with Universal Dependencies}
Part-of-Speech tagging is an essential application within NLP. It is used in machine translation, word meaning disambiguation, parsing, among other applications \citep{chiche2022part}.
It is a highly researched task for which there are many annotated datasets, in many languages. Universal Dependencies \citep[UD;][]{10.1162/coli_a_00402} is a collection of treebanks which include POS tags for hundreds of languages, with a unified POS tagging scheme. UD distinguishes between 17 different tags, called the Universal POS tags \citep{11234/1-5287}. This unified annotation scheme allows the development and comparative evaluation of POS taggers across the languages included in UD. In addition, UD makes it possible to build multilingual POS taggers and dependency parsers which can deal with multiple languages within a single model \citep[e.g.][]{kondratyuk-straka-2019-75}, and this enables cross-lingual transfer.
A limitation of UD is that the annotation quality varies considerably across treebanks \citep{kulmizev-nivre-2023-investigating}. This may negatively impact cross-lingual transfer, a question we investigate in this paper.

\subsection*{Zero-shot learning}
A zero-shot model is a learning model that can perform a task without having seen examples or data of that task during the fine-tuning phase. In the context of this paper, a zero-shot POS tagging model refers to a model that is trained to POS tag sentences in one or more support training languages. The performance of this model is then evaluated on data from a different language, also known as the target language. Importantly, the model does not encounter the data of this target language during the fine-tuning phase (although it may have seen target language data during pre-training).
One could compare it to a student who has an Afrikaans exam scheduled but is only allowed to prepare by studying, for instance, Dutch and German. This approach is useful in NLP to fill gaps in the availability or correctness of data for a target language. Thus, for \textit{extremely low-resource} languages, or especially languages for which no annotated data is available, POS tagging could be performed using a zero-shot model trained on related languages.

We use the zero-shot strategy here because we are interested in scenarios where no data is available for certain languages. This provides a better understanding of the situation because all models developed for this purpose would, by definition, be zero-shot.

\subsection*{Low-resource languages}
While there is no general definition of the term low-resource, researchers have attempted to define it \citep{joshi-etal-2020-state}. However, this definition has not yet been widely adopted. We consider a dataset, and thus a language, to be low-resource if it contains fewer than 50,000 tokens in UD. This mainly concerns indigenous languages, but can also include languages that are more broadly used.
In UD, a token is a syntactic word used for analysis, which might differ from orthographic or phonological words \citep{10.1162/coli_a_00402}.

\section{Methodology}
\label{sec:metho}
Through our experiments, we hope to gain insights into the characteristics of datasets that contribute to the performance of zero-shot POS tagging models for low-resource languages.

Our experiments focus on fine-tuning POS tagging models based on mBERT, a \textit{large language model} \citep[LLM;][]{devlin-etal-2019-bert}. This process involves fine-tuning the model on annotated treebanks to enable it to perform POS tagging. The model's performance is then evaluated on a test dataset. We specifically chose mBERT because of its multilingual capabilities. Additionally, mBERT has shown good results in zero-shot scenarios \citep{pires-etal-2019-multilingual}.

The fine-tuning of mBERT is done using the tool MaChAmp \citep{van-der-goot-etal-2021-massive}. MaChAmp is a user-friendly tool that enables the fine-tuning of LLMs on various NLP tasks from diverse datasets and languages. The latter functionality in particular is valuable for our research. This enables us to jointly train POS tagging models on multiple languages.

\subsection*{Languages and treebanks}
\label{sec:langs}
\begin{table}[]
    \centering
    \begin{tabular}{l l l l}
        \textbf{Language} & \textbf{Treebank} & \textbf{\# Sents} & \textbf{Rank}\\
        \hline
        Afrikaans & AfriBooms & 1.5k & 46 \\
        \hline
        Dutch & Alpino & 12k & 29 \\
        German & GSD & 15k & 19 \\
        English & EWT & 16k & 23 \\
        \hline\hline
        Faroese & FarPaHC & 1.6k & 62\\
        \hline
        Icelandic & Modern & 3.5k & 34\\
        Danish & DDT & 5.5k & 53\\
        Norwegian & Bokmaal & 20k & 14\\
        Swedish & Talbanken & 6.0k & 57\\
        \hline\hline
        Upper Sorbian & UFAL & 0.6k\\
        \hline
        Czech & FicTree & 12.7k & 24\\
        Polish & LFG & 17.2k & 20\\
        Slovak & SNK & 10.6k & 42\\
        \hline
    \end{tabular}
    \caption{Languages with corresponding treebanks (UD v2.13) ranked according to their relatedness to the target language, including the number of sentences (\textbf{\# Sents}) in each treebank and their respective quality rankings \citep[\textbf{Rank};][]{kulmizev-nivre-2023-investigating}.}
    \label{tab:first}
\end{table}

We select three different groups from Universal Dependencies v2.13 \citep{11234/1-5287}. These groups are shown in \autoref{tab:first}. For each group, we select one low-resource language and a few related languages for which larger treebanks are available.

Our focus is primarily on Afrikaans. For this, we use the AfriBooms treebank \citep{augustinus-etal-2016-afribooms}. The supporting languages are Dutch, German, and English. To verify our observations of Afrikaans, we use two other clusters. One cluster includes Scandinavian languages, with Faroese as the low-resource language. The related languages are Icelandic, Danish, Norwegian, and Swedish. We also use a West Slavic cluster, with Upper Sorbian as the low-resource language, supported by Czech, Polish, and Slovak.

\subsection*{Characteristics}
To answer \textbf{RQ1} (\emph{What are the essential characteristics of datasets for effectively fine-tuning zero-shot POS tagging models for low-resource languages?)}, we consider two main characteristics: the linguistic relatedness between languages and the quality of the treebank. 
The relevance of the linguistic relatedness is already evident from previous work (see \autoref{sec:related}), but the treebank quality has not been taken into consideration before in spite of being a clear differentiating factor between UD treebanks (see section~\ref{sec:background} \& \ref{sec:related}). Firstly, we take a look at the linguistic relatedness. The support languages are consistently chosen to be of the same genus as the target language. This results in an intrinsic relatedness. In the first cluster, Dutch shows the closest relatedness to Afrikaans \citep{van-zaanen-etal-2014-development}. In the Scandinavian cluster, Faroese is most closely related to Icelandic \citep{snaebjarnarson-etal-2023-transfer}. In the West Slavic cluster, Upper Sorbian is most closely related to Czech \citep{Howson_2017}, followed by Polish.

For the second characteristic of the treebanks, we rely on a ranking developed by \citet{kulmizev-nivre-2023-investigating}. This ranking is based on three criteria: how difficult or easy the treebanks are to parse, how much information they contain that is actually usable by a parser, and how sample efficient they are. We report the rank of the languages considered in this work in \autoref{tab:first}. The supporting languages for Afrikaans, for example, can be ranked as follows: German $>$ English $>$ Dutch. Throughout this paper, we refer to this as the `quality' of a treebank.

As an additional characteristic, we use the size of the dataset. With this, we investigate whether overfitting might occur and determine the optimal number of sentences a model should use.

\subsection*{Experimental setup}
We train zero-shot models for each of the \hyperref[sec:langs]{three clusters}. We fine-tune these models in several ways. First, we fine-tune separate zero-shot models for each distinct supporting language. Then, we fine-tune models based on different combinations of these languages. We repeat this process for the different clusters.

We conduct learning curve experiments to display the performance of the models as they are fine-tuned on increasingly more data. The fine-tuning starts with five sentences and gradually reaches the maximum available number of sentences from the treebanks. For each cluster and each model within the cluster, we repeat this process three times. During each iteration, the sentences of the training dataset of each language are shuffled. This allows for random selection of sentences, which is crucial for ensuring generalisability.

We determine the accuracy of each model using the F1-score, a common metric for assessing the performance of classifiers such as POS tagging models \citep{10.5555/1214993}. The results of these experiments and the associated observations are discussed in the following section.

Finally, we also look at monolingual non-zero-shot models that are fine-tuned on the respective target languages themselves. This means that we fine-tune three distinct models, each using only Afrikaans, Faroese, or Upper Sorbian.
This can then provide us with an answer to our second question (\textbf{RQ2}): \emph{Are zero-shot models useful in realistic low-resource settings compared to monolingual non-zero-shot models in terms of accuracy?}

\section{Results and discussion}
First, we analyse the learning curves of all clusters to identify which characteristics of datasets seem to have the most impact on the results and could therefore be more suitable for a zero-shot model. Then, we evaluate the effective usability of our models and results to determine whether the zero-shot approach is effective.

\subsection{RQ1: Dataset characteristics}

\subsubsection*{Afrikaans}
\begin{figure}[t]
  \includegraphics[width=\columnwidth]{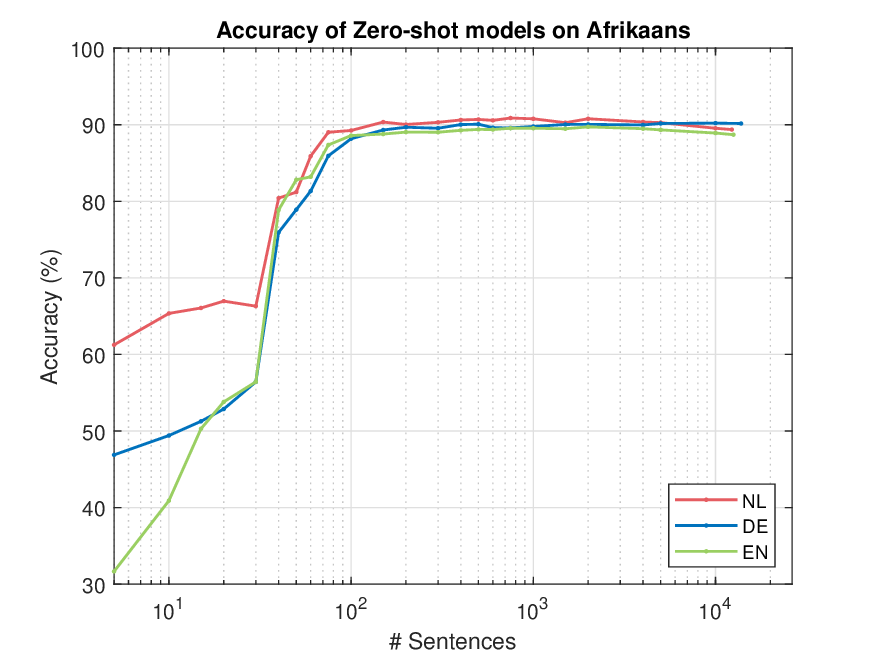}
  \caption{Accuracy of fine-tuned models on Afrikaans, represented through learning curves.}
    \label{fig:first}
\end{figure}

First and foremost, we take a look at the accuracies of the models that have Afrikaans as the target language and that were trained on one supporting language. This can be seen in \autoref{fig:first}. It can be clearly seen that when fine-tuning the zero-shot model using a language more closely related to the target language, the initial accuracies are higher. Initial accuracies are the accuracies that occur with a smaller number of training sentences. A one-to-one correspondence can be seen between how closely related the training language is to the target language and how accurate the corresponding model is for a very small subset of the dataset.

As more sentences are added to the training dataset of the models, it can be seen that the accuracies of the three models converge. This is most likely due to the fact that the supporting languages are all Germanic languages and there is a high similarity to the target language. However, it can be seen that the Dutch model performs better overall and also achieves the best accuracy of the three models. The model that performs the worst overall is the English model. This is not surprising, as English is least closely related to Afrikaans among the three supporting languages, and its quality falls in the middle range.

\begin{figure}[t]
  \includegraphics[width=\columnwidth]{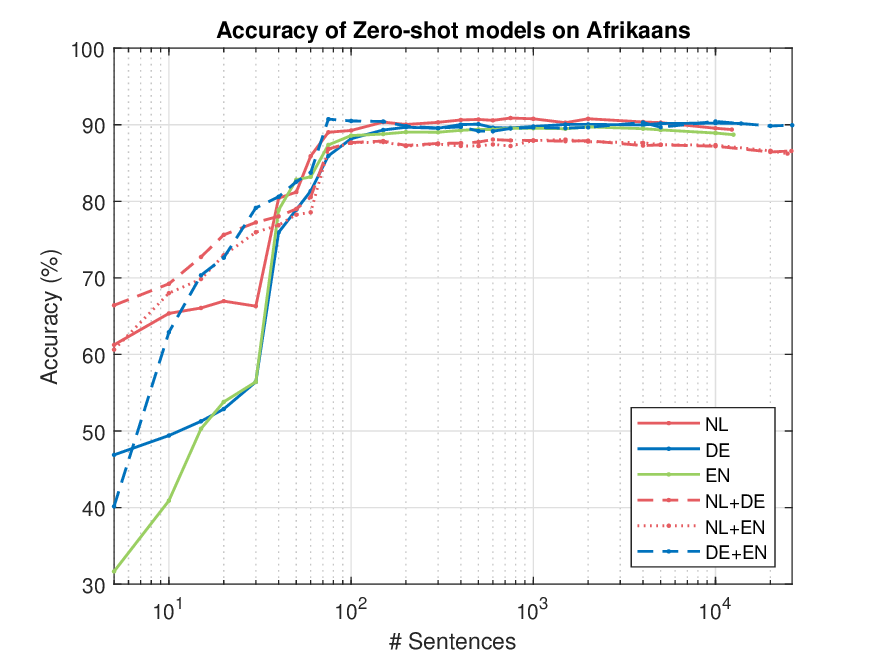}
  \caption{Accuracy of fine-tuned models on Afrikaans, represented through learning curves.}
    \label{fig:second}
\end{figure}

Next, we add the models that were trained with combinations of supporting languages and analyse these separately. This can be seen in \autoref{fig:second}. It is immediately noticeable that the added models using the most closely related language initially achieve higher accuracies. The model with the highest initial accuracy here is the Dutch-German model. This seems logical since Dutch and German follow each other as most related to Afrikaans. In second place is the Dutch-English model, again because Dutch is in the training dataset and has a significant influence. The worst-performing model is the German-English model. This is not surprising since German and English are the two languages that are least related to Afrikaans. 

As the size of the training datasets of the models increases, there is a greater shift between the accuracies of the different models. The model that performs best overall is the German-English model. This is unexpected, given that Dutch is the closest language to Afrikaans \citep{doi:10.5842/47-0-649}. One explanation for this might be that when a model is trained on more than one language, the quality of the datasets becomes more important than the language relatedness. The Dutch-German model performs slightly better across the board than the Dutch-English model, which suggests that relatedness still plays a role. When we look at all the models in \autoref{fig:second} globally, we can make several observations:

If a model has a language in the training dataset that is more closely related to the target language, the model has a higher initial accuracy. When multiple closely related languages are used, such as the Dutch-German model, this accuracy increases even further.

The most performant model is one trained on multiple languages. In the case of Afrikaans, this is the German-English model. This can be attributed to the quality of the datasets used. This model quickly achieves better results and consistently maintains a high accuracy.

There seems to be a plateau at which all models achieve accuracies that neither increase nor decrease, usually between 100 and 5000 sentences. What is also notable is that within this interval, the Dutch model generally performs the best, while the German-English model achieved the highest peak accuracy prior to this interval.

\subsubsection*{Faroese}

\begin{figure}[t]
  \includegraphics[width=\columnwidth]{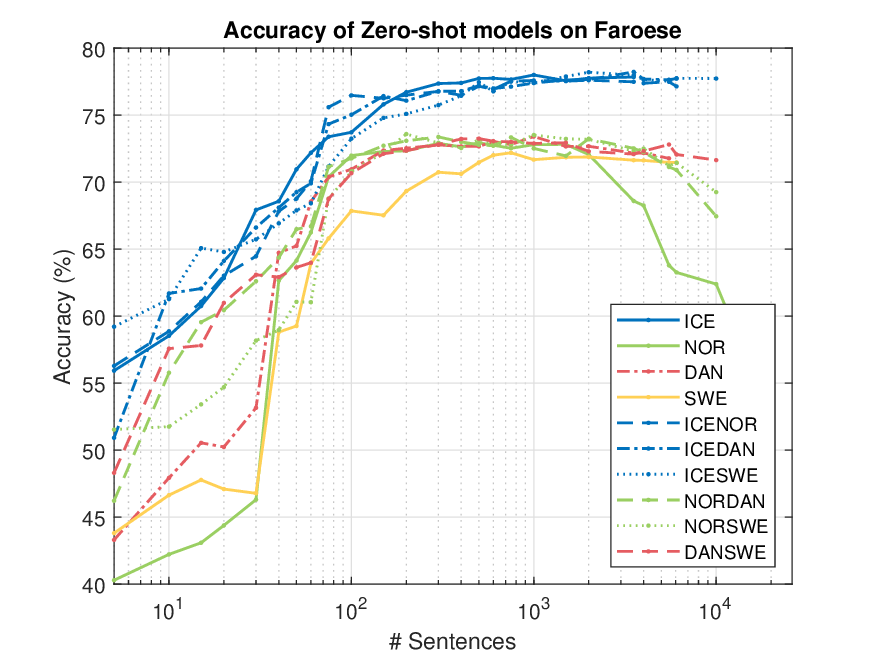}
  \caption{Accuracy of fine-tuned models on Faroese, represented through learning curves.}
    \label{fig:third}
\end{figure}

Secondly, we take a look at all the models we have fine-tuned that have Faroese as the target language. This can be seen in \autoref{fig:third}. What stands out immediately is that all models that contain the most related language - Icelandic - consistently achieve the best results. This results in two distinct groups: one group with models containing Icelandic, and a second group with the other models. Just as with Afrikaans, it can also be seen here that the models that contain a more related language achieve a higher initial accuracy.

Regarding dataset quality, the model fine-tuned using the highest-quality datasets (Icelandic-Norwegian) ranks among the best performing models, while the model trained on the lowest-quality datasets (Danish-Swedish) ranks among the worst. Interestingly, the model that achieves the overall peak accuracy is the Icelandic-Danish model, again highlighting the importance of language relatedness, not only for lower training sizes, but throughout the entire process.

Here, a plateau between about 100 and 5000 sentences can also be clearly seen. This is especially noticeable in the accuracies of the models that do not include Icelandic. The curves start with an increase, followed by a stagnation and then a decline in accuracy. This is particularly noticeable in the models trained on Swedish and Norwegian. In the case of Swedish, this is not surprising as the treebank is of relatively low quality and not very closely related to Faroese. Norwegian, on the other hand, is closely related and of high quality, which makes this trend all the more striking.

Some further observations also become clear here. As more sentences are added to the training dataset of the models, the accuracies of the models converge. This again highlights the idea of intrinsic relatedness between languages within the same language family. Furthermore, it can be seen that the best performing model is one that is fine-tuned on multiple supporting languages, although the model solely fine-tuned on Icelandic is also among the better performing models.

\subsubsection*{Upper Sorbian}

\begin{figure}[t]
  \includegraphics[width=\columnwidth]{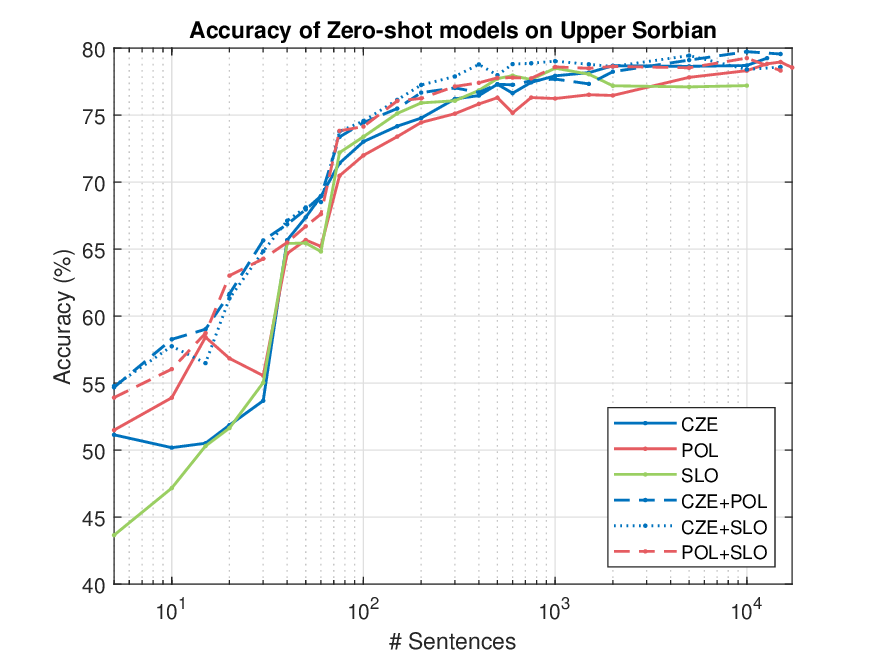}
  \caption{Accuracy of fine-tuned models on Upper Sorbian, represented through learning curves.}
    \label{fig:fourth}
\end{figure}

Lastly, we look at the accuracies of models whose target language is Upper Sorbian. This can be seen in \autoref{fig:fourth}. Here, largely the same trends are seen as in the two previous clusters. The initial accuracies of models trained with the most closely related language (Czech) are higher, although Polish takes the lead when the model is trained on a single supporting language. The statement holds true for models trained on multiple supporting languages: the greater the relatedness and the higher the quality, the better the model performs. In addition, the trend between 100 and 5000 sentences can be seen again here, although it is slightly less pronounced.

\subsection{RQ2: Usefulness of zero-shot models in realistic settings}

\begin{table}[]
    \centering
    \begin{tabular}{ccc}
        \textbf{Language} & \textbf{Zero-shot} & \textbf{Non-zero-shot} \\
        \hline
        Afrikaans & 90.9 & \textbf{98.5} \\
        Faroese & 78.2 & \textbf{97.3} \\
        Upper Sorbian & \textbf{79.7} & 70.7 \\
        \hline
    \end{tabular}
    \caption{Peak accuracies of zero-shot \& non-zero-shot models.}
    \label{tab:second}
\end{table}
In order to answer \textbf{RQ2}, we compare our zero-shot models with those that have been fine-tuned directly on the respective target languages (non-zero-shot).

\subsubsection*{Comparing zero-shot and non-zero-shot performance}

Firstly, we examine the practical relevance of zero-shot models in the context of low-resource languages. The peak accuracies for both our zero-shot and non-zero-shot models can be seen in \autoref{tab:second}. For Afrikaans and Faroese, we observe that the non-zero-shot models outperform their zero-shot counterparts, with the Faroese model showing a nearly 20 percentage point improvement over the zero-shot model. This suggests that, given enough training data, fine-tuning on the target language can lead to substantially better results, as also discussed by \citet{meechan-maddon-nivre-2019-parse}.

However, when we take a look at Upper Sorbian, an \textit{extremely low-resource} language with only 23 training sentences, we observe a different trend. Here, the zero-shot model actually surpasses the non-zero-shot model by 9 percentage points, achieving a peak accuracy of 79.7\% compared to the non-zero-shot model's 70.7\%. This result suggests that our zero-shot models are certainly a viable option for extremely low-resource languages or languages for which no data is available.

Upper Sorbian is not a unique case; we counted 82 languages in UD v2.14 that have fewer than 23 training sentences. This widespread scarcity highlights the importance of zero-shot models in real-world applications where data is often hard to come by.

\subsubsection*{Amount of annotated data needed to surpass zero-shot performance}

\begin{figure}[t]
  \includegraphics[width=\columnwidth]{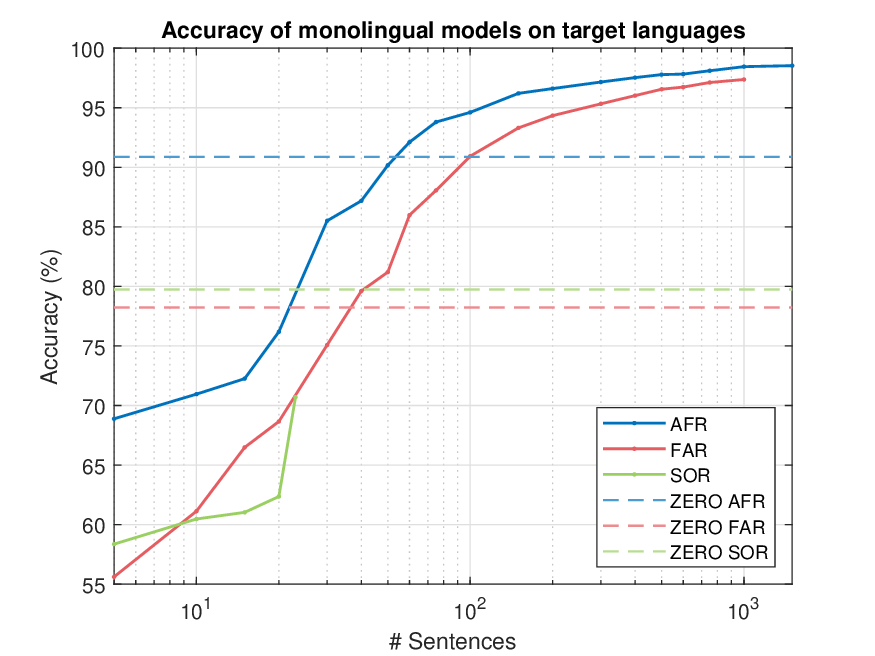}
  \caption{Accuracy of models fine-tuned on the target languages Afrikaans, Faroese, and Upper Sorbian, represented through learning curves accompanied by the peak accuracies of the respective zero-shot models.}
    \label{fig:fifth}
\end{figure}

Secondly, we take a look at how much annotated data is necessary to improve upon zero-shot performance through monolingual fine-tuning (non-zero-shot). In \autoref{fig:fifth}, the accuracies of the POS tagging models are shown when they are fine-tuned on the respective target languages, alongside the peak accuracies of the respective zero-shot models. Using these lines, and the intersection they make with the learning curves of the non-zero-shot models, we can estimate when a non-zero-shot model becomes strictly better than a zero-shot model for the same target language.

For Afrikaans, the intersection occurs between 50 and 60 training sentences, indicating that at least this amount is necessary for the non-zero-shot model to outperform the zero-shot model. Similarly, for Faroese, the intersection point is around 40 sentences, suggesting a slightly lower data requirement to achieve better performance. Again, the Upper Sorbian models are slightly different. The learning curve for the non-zero-shot model does not intersect with the peak accuracy of the zero-shot model, simply because there is not enough targeted data available.

\citet{meechan-maddon-nivre-2019-parse} made similar observations in a similar context. They compared dependency parsers trained on treebanks from several support languages (akin to our zero-shot setting) with those trained solely on target language data (similar to our non-zero-shot setting). They found that the non-zero-shot models required between 100 and 200 sentences to reach the performance of the zero-shot models. This is higher than what we found, but can likely be explained by the fact that they do not make any use of pretraining and use a BiLSTM instead of a transformer. We use mBERT, which has a greater number of parameters than a BiLSTM and which already has acquired cross-lingual transfer capabilities by virtue of being trained on multilingual data. This may reduce the need for annotated target language data. This should be verified in future research, however. 

\section{Related work}
\label{sec:related}
Dealing with the limited availability of training data for low-resource languages is an active area of research within NLP.
Thanks to the UD treebanks, a large collection of data in numerous languages with varying data sizes, POS tagging and dependency parsing have become highly researched topics within this context

Closest to our work, \citet{de-vries-etal-2021-adapting} investigated zero-shot transfer for two target languages: Gronings and West Frisian. They also fine-tuned mBERT on related languages, as well as monolingual language models in related languages. They found the latter to be superior to the former.
Relatedly, \citet{de-vries-etal-2022-make} did an extensive evaluation of zero-shot POS tagging across 105 target languages. They fine-tuned mBERT using 65 different support languages, testing all possible combinations of support and target languages, with one support language used each time. They found that related languages are generally the best support languages.

Our work is complementary to these by considering a number of target languages that is in between these two extremes (2 versus 105). It allows a targeted evaluation, looking at learning curves and trying multiple support languages in different combinations, while still providing results that generalize to more than two closely related languages. We confirm that, among related languages, the ones that are the most closely related to the target language are the best support languages. This finding is consistent with many other earlier works in POS tagging and dependency parsing using different types of models \citep{smith-etal-2018-82,pires-etal-2019-multilingual,lauscher-etal-2020-zero}.

Our learning curve experiments take inspiration from earlier work in dependency parsing by \citet{meechan-maddon-nivre-2019-parse}. They investigated zero- and few-shot learning of multilingual parsers to find out how much can be gained from cross-lingual transfer versus annotating target language data. They use a BiLSTM parser trained only on treebank data, in multiple languages, including and excluding target language data. Their results showed that the zero-shot approach is inferior to the other approaches, provided at least 200 training sentences are available from the target language. We confirm this finding in the context of fine-tuning a multilingual transformer model, although we find that fewer training sentences are necessary in this context.

Finally, a dataset property which has not yet been investigated in the context of cross-lingual transfer (to our knowledge) is data quality.
\citet{kulmizev-nivre-2023-investigating} thoroughly evaluated the quality of UD treebanks using three different metrics and found that the quality varies considerably across treebanks. They found some treebanks to perform consistently low across metrics, making them practically unusable. This raises the question of how this quality impacts results in cross-lingual transfer: a low-quality treebank may be too noisy to use for cross-lingual transfer. We investigated this question and found a subtle link between the quality of the UD Treebanks and the peak accuracies of the corresponding zero-shot models. Of course, more research is needed to confirm this by investigating a larger set of treebanks.

\section{Conclusion}
\label{sec:conclusion}
Initially, we can conclude that developing zero-shot POS tagging models is a viable option for low-resource languages. Nevertheless, using the low-resource dataset of a specific language remains superior for constructing a POS tagging model for that language, similar to what \citet{meechan-maddon-nivre-2019-parse} found in the context of dependency parsing. 
If the amount of data for a language is so scarce and/or a zero-shot model is still desired, the following guidelines can be followed:

\textit{One language can be used as a support language.} In this case, always use the language that is most closely related to the target language. This generally gives better accuracies with a low number of training sentences. Even with larger numbers of training sentences, these models tend to perform well. The quality plays a lesser role here.

\textit{Multiple support languages can be used.} In this case, use as many languages as possible that are closely related to the target language and are of high quality. High relatedness gives the best results with a limited number of training sentences, and high quality generally gives the best results with higher numbers of training sentences.

\textit{What is the most suitable number of training sentences?} If enough data is available from the support languages, preferably use a training number of 100 to 5000 sentences. Below 100 sentences, the models are often `underfitted'. Above 5000 sentences, the models can overfit and the accuracies may decrease.

\section{Limitations}

This work considered only three target languages, each paired with three to four related source languages, selected somewhat arbitrarily. While this restricted number allowed in-depth analysis, our findings need to be verified using more languages with various degrees of relatedness. To keep the number of languages manageable while ensuring generalizability of the results, a sample of typologically diverse languages could be selected using the recently proposed framework by \citet{ploeger2024principled}.

Additionally, we relied on the linguistic literature to describe the degree of relatedness between the languages considered. It would be informative to quantitatively define language distances, as done by \citet{ploeger2024principled}.

Finally, we only evaluated one pretrained language model, mBERT. Our findings should be verified with other models.

\section*{Acknowledgments}
We want to thank Kushal Tatariya for technical help and Artur Kulmizev for the theoretical insights. We are also grateful to Max Verbinnen and Sander Verwimp for their help and support throughout this study. We also acknowledge the VSC (Vlaams Supercomputer Centrum) for providing the computational resources used in this work.

\bibliography{low_res_pow}
\bibliographystyle{acl_natbib}
\end{document}